\pdfoutput=1 
\documentclass[10pt,twocolumn,letterpaper]{article}

\usepackage{cvpr}   
\usepackage{graphicx}
\usepackage{subcaption}
\usepackage{caption,subcaption}
\usepackage{amsmath}
\usepackage{amssymb}
\usepackage{amscd}
\usepackage{booktabs}
\usepackage{algorithm}
\usepackage{algpseudocode}
\usepackage{amsthm}
\usepackage{textcomp}
\usepackage{amsthm,amsmath,amssymb}
\usepackage{mathrsfs}
\usepackage{graphicx}
\usepackage{wrapfig}
\usepackage{bbding}
\usepackage{float}
\usepackage{mathrsfs}
\usepackage{bbm}
\usepackage[accsupp]{axessibility}

\newtheorem{theorem}{Theorem}

\usepackage[pagebackref,breaklinks,colorlinks]{hyperref}

\usepackage{lipsum}

\newcommand\blfootnote[1]{%
\begingroup
\renewcommand\thefootnote{}\footnote{#1}%
\addtocounter{footnote}{-1}%
\endgroup
}

\usepackage[capitalize]{cleveref}
\crefname{section}{Sec.}{Secs.}
\Crefname{section}{Section}{Sections}
\Crefname{table}{Table}{Tables}
\crefname{table}{Tab.}{Tabs.}

\begin{document}

\title{Balanced Multimodal Learning via On-the-fly Gradient Modulation}

\author{
\textbf{Xiaokang Peng}\textsuperscript{1,$\dagger$}, 
\textbf{Yake Wei}\textsuperscript{1,$\dagger$}, 
\textbf{Andong Deng}\textsuperscript{2}, 
\textbf{Dong Wang}\textsuperscript{3}, 
\textbf{Di Hu}\textsuperscript{1,*}
\vspace{0.5em}
\\
\textsuperscript{1}Gaoling School of Artificial Intelligence, Renmin University of China, Beijing\\
\textsuperscript{1}Beijing Key Laboratory of Big Data Management and Analysis Methods, Beijing\\
\textsuperscript{2}Shanghai Jiao Tong University, Shanghai
\\
\textsuperscript{3}Shanghai Artificial Intelligence Laboratory, Shanghai
\\
\textsuperscript{1}\{xiaokangpeng, yakewei, dihu\}@ruc.edu.cn, \textsuperscript{2}\{andongdeng69, dongwang.dw93\}@gmail.com
}

\maketitle

\begin{abstract}
    Multimodal learning helps to comprehensively understand the world, by integrating different senses. Accordingly, multiple input modalities are expected to boost model performance, but we actually find that they are not fully exploited even when the multimodal model outperforms its uni-modal counterpart. Specifically, in this paper we point out that existing multimodal discriminative models, in which uniform objective is designed for all modalities, could remain under-optimized uni-modal representations, caused by another dominated modality in some scenarios, e.g., sound in blowing wind event, vision in drawing picture event, etc. To alleviate this optimization imbalance, we propose on-the-fly gradient modulation to adaptively control the optimization of each modality, via monitoring the discrepancy of their contribution towards the learning objective. Further, an extra Gaussian noise that changes dynamically is introduced to avoid possible generalization drop caused by gradient modulation. As a result, we achieve considerable improvement over common fusion methods on different multimodal tasks, and this simple strategy can also boost existing multimodal methods, which illustrates its efficacy and versatility. The source code is available at \url{https://github.com/GeWu-Lab/OGM-GE_CVPR2022}.\blfootnote{\noindent
    \textsuperscript{$\dagger$}Equal contribution. 
    \textsuperscript{*}Corresponding author.
    }
\end{abstract}
\section{Introduction}

\begin{figure*}

\centering
\subcaptionbox{\label{1}}{\includegraphics[width = .33\linewidth]{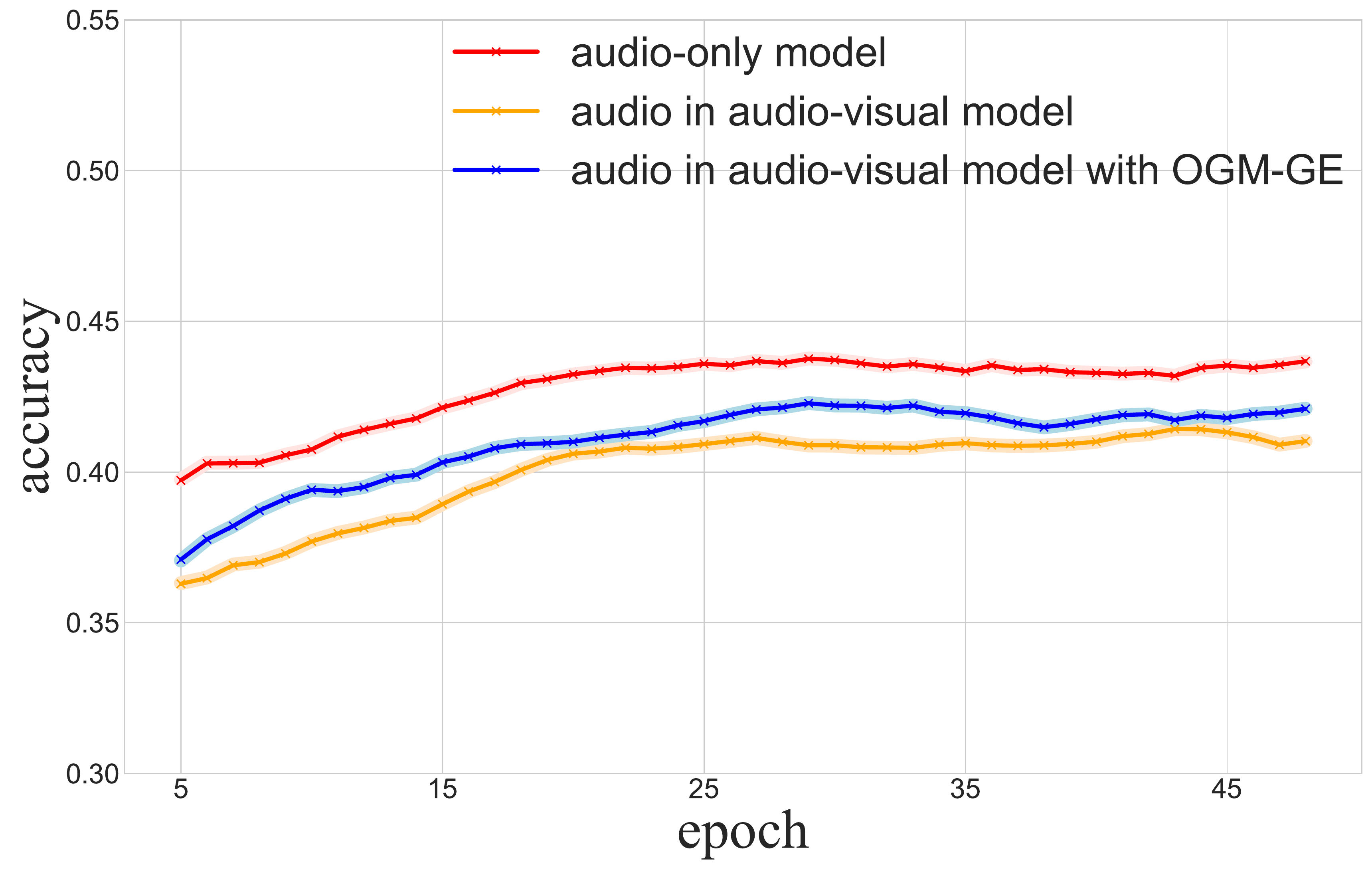}}\hfill
\subcaptionbox{\label{2}}{\includegraphics[width = .33\linewidth]{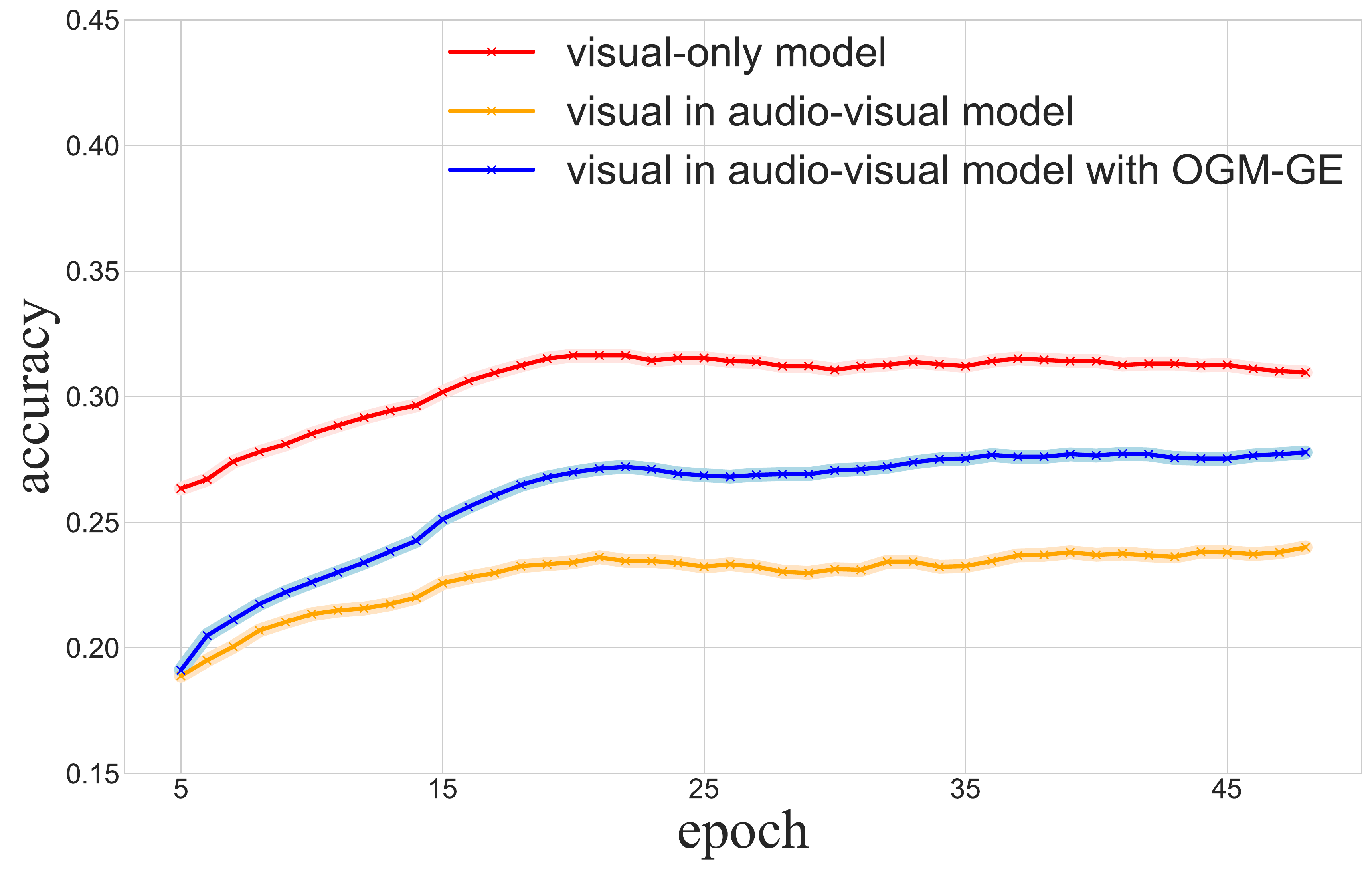}}
\subcaptionbox{\label{2}}{\includegraphics[width = .33\linewidth]{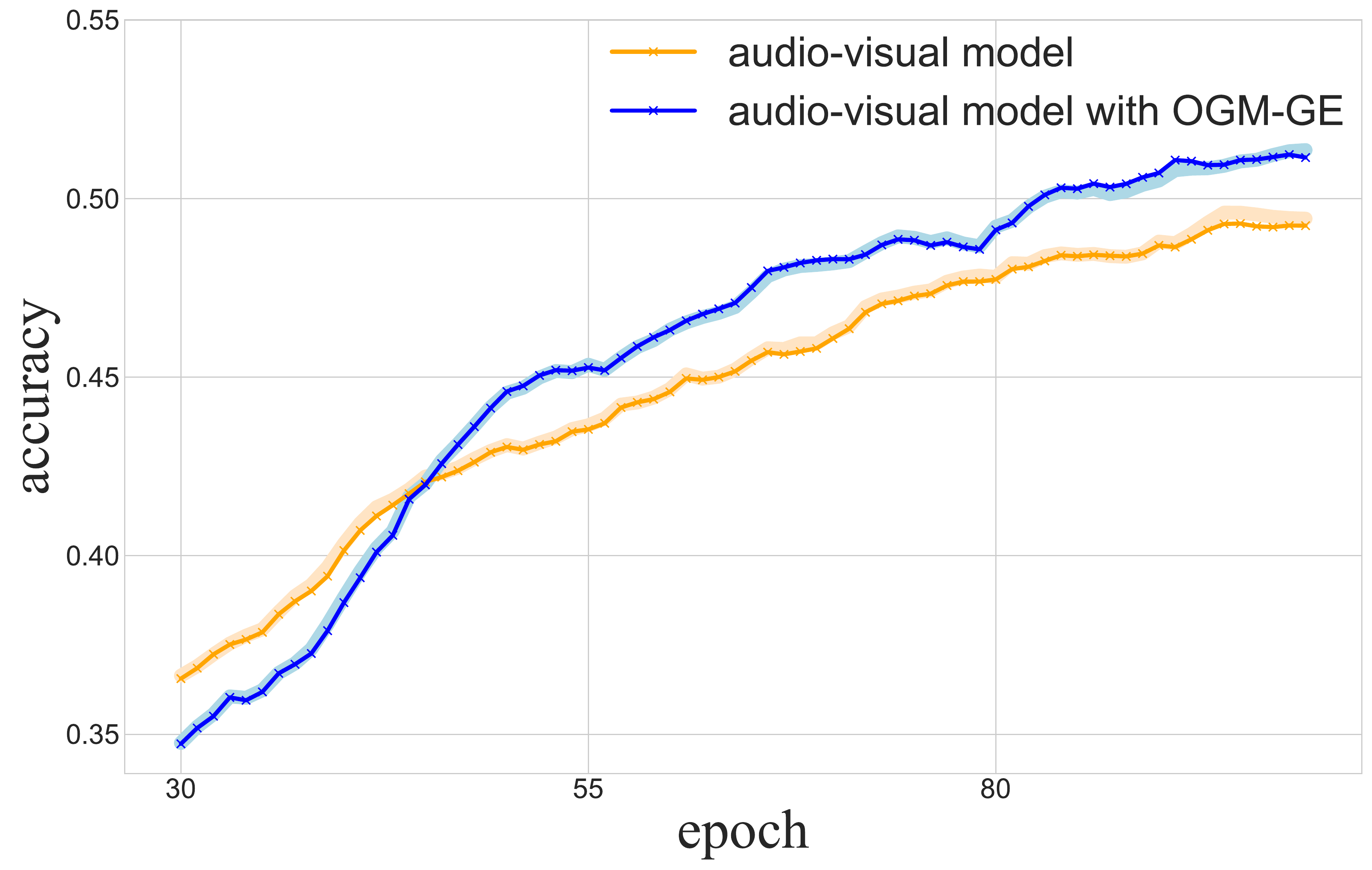}}

\caption{Performance of the uni-modal models, joint-trained multimodal model, and multimodal model with our proposed OGM-GE on the validation set of the VGGSound \cite{chen2020vggsound} dataset. (a) Performance of audio modality. (b) Performance of visual modality. (c) Performance of the multimodal model. Best viewed in color.}

\label{five_lines} 
\end{figure*}

People perceive the world by collaboratively utilizing multiple senses: eyes to look, ears to listen, and hands to touch. Such a multimodal way can provide more comprehensive information from different aspects. Inspired by the multi-sensory integration ability of humans~\cite{gazzaniga2006cognitive}, multimodal data, collected from different sensors, tend to be more considered in machine learning. In recent years, multimodal learning has exhibited a clear advantage in improving the performance of previous uni-modal tasks as well as addressing new challenging problems, such as action recognition~\cite{nagrani2020speech2action,gao2020listen,kazakos2019epic,wang2020temporal}, audio-visual speech recognition~\cite{potamianos2004audio}, and visual question answering~\cite{antol2015vqa,ilievski2017multimodal,wu2021star}.

Multimodal data usually provides more views compared with uni-modal one, accordingly learning with multimodal data should match or outperform the uni-modal case. However, according to the recent study~\cite{wang2020makes}, multimodal models that optimize the uniform learning objective for all modalities with joint training strategy can be inferior to the uni-modal model in some situations. Such a phenomenon violates the intention of improving model performance through integrating information from multiple modalities. Previous researchers claimed that various modalities tend to converge at different rates, leading to uncoordinated convergence problem~\cite{wang2020makes,sun2021learning,ismail2020improving}. To cope with this problem, some methods aid the training of multimodal models with the help of additional uni-modal classifiers or pre-trained models~\cite{wang2020makes,du2021improving}. Hence, they inevitably bring extra efforts on training additional neural modules. 

However, we further find that even when the multimodal models \emph{outperform} the uni-modal ones, they still cannot fully exploit the potential of multiple modalities. As shown in Figure~\ref{five_lines}, the joint multimodal model achieves the best event classification performance on VGGSound~\cite{chen2020vggsound}, but the performance of visual and audio modality within it is clearly worse than that in the visual-only and audio-only model, respectively\footnote{Here, the visual-only and the audio-only are the models that trained with a single modality. To evaluate the uni-modal encoders of the multimodal model, we observe the performance through fixing the joint-trained uni-modal encoder and finetuning a uni-modal classifier only.}. This interesting observation suggests under-optimized representation in both modalities. We consider the reason could be that, in some multimodal scenarios, the dominated modality~\cite{parida2020coordinated} with better performance (e.g., sound of wind blowing, vision of playing football, etc.) will suppress the optimization of the other one. Moreover, as illustrated in Figure~\ref{five_lines}(a) and (b), there is another noticeable observation that accuracy drops more remarkably in visual modality compared with audio case, which is consistent with the fact that VGGSound, as a curated sound-oriented dataset, prefers audio modality, even the sound sources are guaranteed to be visible. Generally speaking, such dataset preferences will lead to one modality is usually dominant, resulting in this phenomenon of optimization imbalance.

Aiming to solve the problem above, we first analyze the imbalance phenomenon from the optimization perspective, and find that the modality with better performance contributes to lower joint discriminative loss then dominates the optimization progress via propagating limited gradient over the other modality, thus leading to the under-optimized situation. 
Then, to ease the situation, we propose to control the optimization process of each modality via the \emph{On-the-fly Gradient Modulation} (OGM) strategy. 
Specifically, the contribution discrepancy between different modalities to the learning objective is dynamically monitored during training progress, which is then exploited to adaptively modulate the gradients, offering more efforts on the under-optimized modality. However, the modulated gradient may lower the intensity of the stochastic gradient noise, which has been proven to have a positive correlation with generalization ability~\cite{jastrzkebski2017three}. Hence, we further introduce extra Gaussian noise that changes dynamically to achieve \emph{Generalization Enhancement} (GE). 
After applying our method of OGM-GE on the multimodal learning task of VGGSound in Figure~\ref{five_lines}, we obtain the consistent performance boost for under-optimized uni-modal representation, i.e., the blue curves in Figure~\ref{five_lines}(a) and (b). More than that, the visual modality gains more improvement. As a result, our method is noticeably superior to the conventional one in the multimodal learning setting, as shown in Figure~\ref{five_lines}(c). To comprehensively demonstrate the effectiveness of OGM-GE, we test it in various multimodal tasks on different datasets, which brings consistent improvements, working with both vanilla fusion strategies and existing multimodal methods.

To summarize, our contributions are as follows:
\begin{itemize} 
\item We find the optimization imbalance phenomenon that the performance of the joint multimodal model is limited due to the under-optimized representations, and then analyze it from the optimization perspective.
\item The OGM-GE method is proposed to solve the optimization imbalance problem by controlling the optimization process of each modality dynamically as well as enhancing generalization ability. 
\item The proposed OGM-GE can be plugged in not only vanilla fusion strategies but also existing multimodal frameworks and brings consistent improvement, indicating its promising versatility.
\vspace{0.2em}
\end{itemize}

\section{Related works}

\subsection{Multimodal learning}

Multimodal learning is a sophisticated learning paradigm in the machine learning community and has been attracting increasing attention because of the growing amount of multimodal data, which naturally contains abundant correlation. There are different research directions according to specific applications. For example, some researchers have explored the correspondence between multimodal data in an unsupervised manner, to learn meaningful representations for downstream tasks~\cite{aytar2016soundnet,alwassel2020self,korbar2018cooperative,hu2019deep,hu2019dense}. Moreover, there is numerous research dedicating to exploiting information of multiple modalities to boost model performance on a certain task compared with uni-modal frameworks, e.g., action recognition~\cite{nagrani2020speech2action,gao2020listen,kazakos2019epic}, audio-visual speech recognition~\cite{potamianos2004audio,hu2016temporal}, and visual question answering~\cite{antol2015vqa,ilievski2017multimodal}. 
However, most multimodal methods that utilize joint training strategy could not fully exploit all modalities and produce under-optimized uni-modal representations, making the performance of multimodal models fail to reach where they are expected, even though they sometimes indeed outperform the uni-modal counterparts.

\subsection{Imbalanced multimodal learning}

The aforementioned defect of previous audio-visual learning methods encourages researchers to explore the reasons behind it. Several studies pointed out that most multimodal learning methods cannot effectively boost the performance even with more information ~\cite{wang2020makes,winterbottom2020modality,sun2021learning,du2021improving}, which is caused by the discrepancy between modalities. Wang et al.~\cite{wang2020makes} found that different modalities have different convergence rates, making the jointly trained multimodal model fail to match or outperform its uni-modal counterpart. Winterbottom et al.~\cite{winterbottom2020modality} demonstrated an inherent bias in TVQA dataset towards the textual subtitle modality. Recently, several methods have emerged trying to solve these problems~\cite {wang2020makes,sun2021learning,du2021improving}. Wang et al.~\cite{wang2020makes} proposed Gradient-Blending to obtain an optimal blending of modalities based on their over-fitting behaviors. Further, Du et al.~\cite{du2021improving} trained the multimodal model by distilling knowledge from the well-trained uni-modal models to strengthen the uni-modal encoders. These methods can indeed bring improvement to a degree, but extra efforts are required to introduce additional neural modules, which complicates the training procedure. In this work, from the perspective of optimization, we tackle this problem by adaptively controlling the optimization of each modality without extra modules.

\subsection{Stochastic gradient noise}

The gradient noise of SGD is considered to have an essential correlation with the generalization ability of the deep models~\cite{zhu2018anisotropic,chaudhari2018stochastic,xie2021artificial,wu2020noisy,he2019control}. Such stochastic gradient noise is introduced by random mini-batch sampling, believed to serve as a regularization and help the model to escape from saddle point or local optimum~\cite{jastrzkebski2017three,chaudhari2018stochastic,xie2021artificial,wu2020noisy}. Zhou et al.~\cite{zhou2019toward} further provided theoretical proof that the stochastic gradient algorithms, in conjunction with proper Gaussian noise, are guaranteed to converge to the global optimum in polynomial time with random initialization. In this work, to enhance the generation ability of the multimodal model, we introduce extra Gaussian noise into the gradient and achieve considerable improvement.

\section{Method}

\subsection{Optimization imbalance analysis}
\label{imbalance_analysis}

\begin{figure*}
    \centering
    \includegraphics[width=17cm]{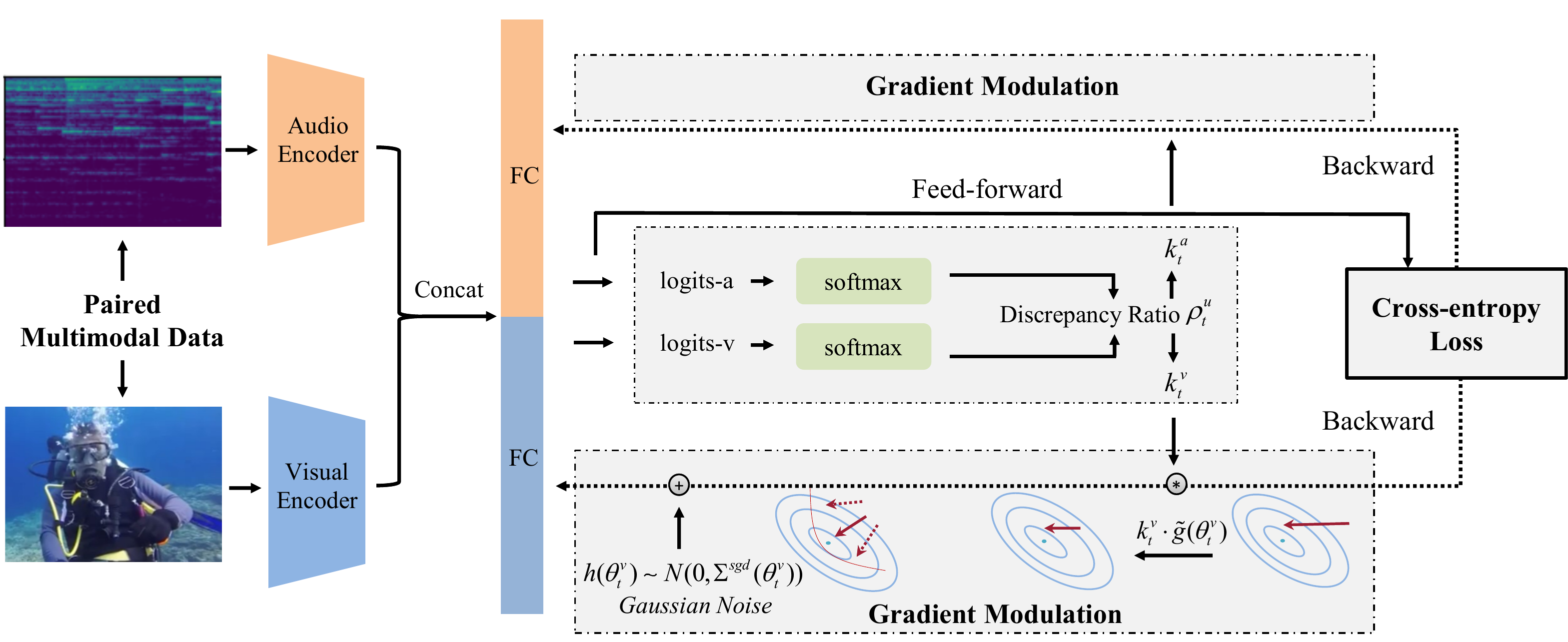}
    \caption{The pipeline of the On-the-fly Gradient Modulation with Generalization Enhancement strategy.}
    \vspace{-0.5em}
    \label{pipeline}
\end{figure*}

We introduce the analysis of the optimization imbalance phenomenon and find that the optimization progress of the multimodal model is dominated by the modality with better performance, leading to another modality being under-optimized. Here we take audio and visual modalities as example. For convenience, we denote training dataset $\mathcal{D}=\{x_{i}, y_{i}\}_{i=1,2...N}$. Each $x_{i}$ consists of two inputs from different modalities as $x_{i}=(x^{a}_{i}, x^{v}_{i})$, where $a$ and $v$ indicate audio and visual modality, respectively. $y_i \in \left\{1,2,\cdots,M\right\}$, where $M$ is the number of categories. We use two encoders $\varphi^{a}(\theta^{a},\cdot)$ and $\varphi^{v}(\theta^{v},\cdot)$ to extract features, where $\theta^{a}$ and $\theta^{v}$ are the parameters of encoders. Representatively, we take the most widely used vanilla fusion method, concatenation, as the example here\footnote{The analysis of another vanilla fusion method, summation, is provided in the \emph{Supp. Materials}. These two simple but effective fusion method are widely used in many existing multimodal method.}. Let $W \in \mathbb{R}^{M \times (d_{\varphi_a}+d_{\varphi_v})}$ and $b \in \mathbb{R}^{M}$ denote the parameters of the last linear classifier. Then the logits output of the multimodal model is as follows:
\begin{equation}
\label{output}
f(x_{i})=W[\varphi^{a}(\theta^{a},x^{a}_{i});\varphi^{v}(\theta^{v},x^{v}_{i})]+b.
\end{equation}

To observe the optimization process of each modality individually, $W$ can be represented as the combination of two blocks: $[W^{a}, W^{v}]$. The Equation~\ref{output} can be rewritten as:
\begin{equation}
\label{output_re}
f(x_{i})=W^{a}\cdot\varphi^{a}(\theta^{a},x^{a}_{i})+ W^{v}\cdot\varphi^{v}(\theta^{v},x^{v}_{i})+b.
\end{equation}

We denote the logits output for class $c$ as $f(x_{i})_{c}$, and the cross-entropy loss of the discriminative model becomes $L=- \frac{1}{N} \sum^{N}_{i=1}  \text{log} \frac{e^{f(x_{i})_{y_i}}}{\sum^M_{k=1}e^{f(x_{i})_{k}}}$ .
With the \emph{Gradient Descent} (GD) optimization method, $W^{a}$ and the parameters of encoder $\varphi^{a}(\theta^{a},\cdot)$ are updated as (similarly for $W^{v}$ and $\varphi^{v}(\theta^{v},\cdot)$ ):
\begin{equation}
\label{update_W}
\begin{aligned}
W_{t+1}^{a} &=W_{t}^{a}-\eta\nabla_{W^{a}} L(W_{t}^{a})\\
&=W_{t}^{a}-\eta \frac{1}{N} \sum^{N}_{i=1} \frac{\partial L }{\partial f(x_{i})} \varphi^{a}(\theta^{a},x^{a}_{i}),
\end{aligned}
\end{equation}
\begin{equation}
\label{update_phi}
\begin{aligned}
\theta_{t+1}^{a} &=\theta_{t}^{a}-\eta\nabla_{\theta^{a}} L(\theta_{t}^{a})\\
&=\theta_{t}^{a}-\eta  \frac{1}{N} \sum^{N}_{i=1} \frac{\partial L }{\partial f(x_{i})}  \frac{\partial( W^{a}_{t}\cdot\varphi^{a}_{t}(\theta^{a},x^{a}_{i}))}{\partial \theta_{t}^{a}},
\end{aligned}
\end{equation}
where $\eta$ is the learning rate.
According to Equation~\ref{update_W} and~\ref{update_phi}, we can find that the optimization of $W^{a}$ and $\varphi^{a}$ nearly has no correlation with that of the other modality\footnote{Our OGM-GE can also boost methods that contain more complex interaction between modalities, please see details in Section~\ref{sec:gene}.}, except the term related to the training loss ($\frac{\partial L }{\partial f(x_{i})}$). The uni-modal encoders thus can hardly make adjustment according to the feedback from each other. Then combined with Equation~\ref{output_re}, the gradient $\frac{\partial L }{\partial f(x_{i})}$ can be rewritten as :
\begin{equation}
\label{ce_bp}
\begin{aligned}
\frac{\partial L }{\partial f(x_{i})}_{c}=
\frac{e^{(W^{a}\cdot\varphi^{a}_i+W^{v}\cdot\varphi^{v}_i+b)_{c}}}{\sum^M_{k=1}e^{(W^{a}\cdot\varphi^{a}_i+W^{v}\cdot\varphi^{v}_i+b)_{k}}} - 1_{c=y_{i}}.
\end{aligned}
\end{equation}

For convenience, we simplify $\varphi^{a}(\theta^{a},x^{a}_{i})$ and $\varphi^{v}(\theta^{v},x^{v}_{i})$ as $\varphi^{a}_i$ and $\varphi^{v}_i$, respectively. Then, we can infer that for sample $x_i$ belonging to class $y_i$, when one modality, such as the visual modality, has better performance, it contributes more to  $\frac{\partial L }{\partial f(x_{i})_{y_i}}$ via $W^{v}\cdot\varphi^{v}_i$, leading to lower loss globally. Accordingly, the audio modality, which relatively has a lower confidence for the correct category, could obtain limited optimization efforts w.r.t. its modal-specific parameters during the back propagation. This phenomenon indicates that the modality with better performance dominates the optimization progress. As a result, when the training of multimodal model is about to converge, the other modality could still suffer from under-optimized representation and needs further training.

\subsection{On-the-fly gradient modulation}
\label{modulation}

As discussed, the optimization progress of multimodal discriminative models is usually dominated by the modality with better performance, leading to the under-optimized representations that limit the model performance. To solve the problem, we aim to amend the optimization process of each modality via OGM strategy, as shown in Figure~\ref{pipeline}.

In this section, we follow the notation in Section~\ref{imbalance_analysis}. The parameters $\theta^{u}$ of the encoder $\varphi^{u}$, where $u \in \{a,v\}$, is updated as follows when using GD method:
\begin{equation}
\label{theta_gd}
\theta_{t+1}^{u} =\theta_{t}^{u}-\eta \nabla_{\theta^{u}} L(\theta_{t}^{u}).
\end{equation}

In practice, we employ the widely used \emph{Stochastic Gradient Descent} (SGD) optimization method and the parameters are updated as:
\begin{equation}
\label{theta_sgd}
\theta^{u}_{t+1} =\theta_{t}^{u}-\eta \tilde{g}(\theta_{t}^{u}),
\end{equation}
where $\tilde{g}(\theta_{t}^{u})=\frac{1}{m}\sum_{x\in  B_{t}}\nabla_{\theta^{u}} \ell (x;\theta_{t}^{u})$ is an unbiased estimation of the full gradient $\nabla_{\theta^{u}} L(\theta_{t}^{u})$. $B_{t}$ is a random mini-batch which is chosen in the $t$-th step with size $m$, and $\nabla_{\theta^{u}} \ell (x;\theta_{t}^{u})$ is the gradient w.r.t. $B_{t}$.

Specific to the optimization imbalance problem discussed in Section~\ref{imbalance_analysis}, we propose to adaptively modulate the gradient of each modality via monitoring the discrepancy of their contribution to the learning objective. Here we design the discrepancy ratio $\rho_t^u$:

\begin{equation}
\label{8}
\begin{gathered}
s_{i}^{a}=  \sum_{k=1}^M 1_{k=y_{i}} \cdot  \text{softmax} (W^{a}_{t} \cdot \varphi^{a}_{t}(\theta^{a},x_{i}^{a})+\frac{b}{2})_{k},\\
s_{i}^{v}=  \sum_{k=1}^M 1_{k=y_{i}} \cdot  \text{softmax} (W^{v}_{t} \cdot \varphi^{v}_{t}(\theta^{v},x_{i}^{v})+\frac{b}{2})_{k},
\end{gathered}
\end{equation}
\begin{equation}
\label{calcu_ratio}
\rho^{v}_{t}=\frac{\sum_{i \in B_{t}}  s_{i}^{v} } {\sum_{i \in B_{t}}  s^{a}_i}.
\end{equation}

$\rho^{a}_{t}$ is accordingly defined as the reciprocal of $\rho^{v}_{t}$. Here we use $(W^{u}_{i} \cdot \varphi_{i}(\theta^{u},x_{i}^{u})+\frac{b}{2})$ as the approximated prediction of modality $u$ to estimate uni-modal performance of the multimodal model\footnote{Inspired by Deep Boltzmann Machine~\cite{salakhutdinov2009deep} that uses the average of the bottom-up weight and the top-down weight to obtain the posterior distribution, we use $\frac{b}{2}$ as the bias term to estimate the uni-modal prediction of the multimodal model.}. With $\rho_t^u$ to dynamically monitor the contribution discrepancy between audio and visual modalities, we are able to adaptively modulate the gradient via:
\begin{equation}
\label{k}
k^{u}_{t}=\left\{\begin{array}{cl}
1-\tanh (\alpha \cdot \rho^{u}_{t}) & \text { }\rho ^{u}_{t}>1 \\
1 & \text { others, }
\end{array}\right.
\end{equation}
where $\alpha$ is a hyper-parameter to control the degree of modulation. We integrate the coefficient $k_t^u$ into SGD optimization method, and $\theta_{t}^{u}$ in iteration $t$ is updated as follows:
\begin{equation}
\label{update_1}
\theta^{u}_{t+1} =\theta_{t}^{u}-\eta \cdot k_{t}^{u}\tilde{g}(\theta_{t}^{u}).
\end{equation}

By means of $k_t^u$, the optimization of modality with better performance ($\rho ^{u}_{t}>1$) is mitigated, while the other modality is not affected and able to get rid of limited optimization efforts, gaining adequate training. Using the SGD optimization method with OGM strategy, the optimization process of each modality is controlled respectively and the imbalance problem can be alleviated.

\subsection{Generalization enhancement}
\label{noise}

In the SGD optimization method, according to the central limit theorem, the gradient $\tilde{g}(\theta_{t}^{u})$ in Equation~\ref{theta_sgd} follows a Gaussian distribution when the batch size $m$ is large enough~\cite{mandt2017stochastic}, i.e.,

\begin{equation}
\label{gauss}
\tilde{g}(\theta_{t}^{u}) \sim  \mathcal{N} (\nabla_{\theta^{u}} L(\theta_{t}^{u}), \Sigma  ^{sgd}(\theta_{t}^{u})),
\end{equation}
\begin{equation}
\label{sgn}
\begin{aligned}
\Sigma ^{sgd}(\theta_{t}^{u}) &\approx \frac{1}{m}[\frac{1}{N} \sum_{i=1}^{N} \nabla_{\theta^{u}}  \ell(x_{i};\theta^{u}_t) \nabla_{\theta^{u}}  \ell(x_{i};\theta^{u}_t)^\mathrm{T}\\ 
&- \nabla_{\theta^{u}} L(\theta_{t}^{u}) \nabla_{\theta^{u}} L(\theta_{t}^{u})^\mathrm{T}].
\end{aligned}
\end{equation}

Then Equation~\ref{theta_sgd} can be rewritten as follows, where $\xi_{t}$ is the noise term:
\begin{equation}
\label{7}
\theta_{t+1}^{u} = \theta_{t}^{u}-\eta \nabla_{\theta^{u}} L(\theta_{t}^{u})+\eta \xi_{t}, \xi_{t} \sim  \mathcal{N}(0,\Sigma^{sgd}(\theta_{t}^{u})).
\end{equation}

\begin{theorem}[SGD generalization ability]
\label{noise_gene}
Noise in SGD is closely related to its generalization ability and larger SGD noise often leads to better generalization~\cite{jastrzkebski2017three}. The covariance of the SGD noise is proportional to the ratio of learning rate to batch size.
\end{theorem}

According to Theorem~\ref{noise_gene}, higher value of the gradient covariance ($\xi_{t}$ in Equation~\ref{7}) often brings better generalization ability. When we use the coefficient $k_{t}^{u}$ to modulate the gradient, $\theta^{u}$ is updated as:
\begin{equation}
\label{10}
\begin{gathered}
\theta_{t+1}^{u} =\theta_{t}^{u}-\eta \nabla_{\theta^{u}} L^{\prime}(\theta_{t}^{u})+\eta \xi_{t}^{\prime}, \\
\xi_{t}^{\prime} \sim  \mathcal{N}(0,(k_{t}^{u})^{2}\cdot\Sigma^{sgd}(\theta_{t}^{u})),
\end{gathered}
\end{equation}
where $\eta \nabla_{\theta^{u}} L^{\prime}(\theta_{t}^{u})=k_{t}^{u} \cdot \eta \nabla_{\theta^{u}} L(\theta_{t}^{u})$. Based on Equation~\ref{k}, $k_{t}^{u} \in (0,1]$. Absolutely, when the learning rate and batch-size are fixed, the covariance of $\xi_{t}^{\prime}$ is smaller than the original $\xi_{t}$. The generalization ability of SGD optimization method could be reduced. Hence, it is desirable to develop a method of controlling the SGD noise, recovering the generalization ability.

To enhance the SGD noise, we introduce a simple but effective \emph{Generalization Enhancement} (GE) method that adds a randomly sampled Gaussian noise $h(\theta_{t}^{u}) \sim  \mathcal{N} (0, \Sigma  ^{sgd}(\theta_{t}^{u}))$ to the gradient, which owns the same covariance as $\tilde{g}(\theta_{t}^{u})$ and changes dynamically according to the current iteration. Then, combining the OGM method, $\theta_{t}^{u}$ is updated as:

\begin{equation}
\label{update_n}
\begin{aligned}
\theta^{u}_{t+1} &=\theta_{t}^{u} - \eta (k_{t}^{u}\tilde{g}(\theta_{t}^{u})+ h(\theta_{t}^{u}))\\
&=\theta_{t}^{u}-\eta \nabla_{\theta^{u}} L^{\prime}(\theta_{t}^{u})+\eta \xi_{t}^{\prime}+ \eta \epsilon _{t},
\end{aligned}
\end{equation}
where $\epsilon _{t} \sim  \mathcal{N} (0, \Sigma  ^{sgd}(\theta_{t}^{u}))$. Since $\epsilon _{t} $ and $\xi_{t} $ are independent, Equation~\ref{update_n} can be rewritten as:

\begin{equation}
\label{update_final}
\begin{gathered}
\theta^{u}_{t+1} =\theta_{t}^{u}-\eta \nabla_{\theta^{u}} L^{\prime}(\theta_{t}^{u})+\eta \xi_{t}^{\prime\prime},\\
\xi_{t}^{\prime\prime} \sim  \mathcal{N} (0, ((k_{t}^{u})^{2}+1)\Sigma  ^{sgd}(\theta_{t}^{u})).
\end{gathered}
\end{equation}

Hence, the covariance of the SGD noise is recovered and even enhanced. We provide the overall OGM-GE strategy in Algorithm~\ref{alg:1}, and the pipeline is shown in Figure~\ref{pipeline}. Combining with our strategy, the optimization of multimodal model can be more balanced with guaranteed generalization ability for each modality.

\begin{algorithm}
\caption{Multimodal learning with OGM-GE strategy}
\label{alg:1}
\begin{algorithmic}
\Require Training dataset $\mathcal{D}=\{(x^a_{i},x^v_{i}), y_{i}\}_{i=1,2...N}$, iteration number $T$, hyper-parameter $\alpha$, initialized modal-specific parameters $\theta^{u}$, $u \in \{a,v\}$.
\For{$t=0,\cdots,T-1$} 
    \State Sample a fresh mini-batch $B_{t}$ from $\mathcal{D}$;
    \State Feed-forward the batched data $B_{t}$ to the model;
    \State Calculate $\rho^{u}$ using Equation~\ref{8} and~\ref{calcu_ratio};
    \State Calculate $k_{t}^{u}$ using Equation~\ref{k};
    \State Calculate gradient $\tilde{g}(\theta_{t}^{u})$ using back-propagation;
    \State Sample $h(\theta_{t}^{u})$ based on covariance of gradient $\tilde{g}(\theta_{t}^{u})$;
    \State Update using $\theta^{u}_{t+1} =\theta_{t}^{u} - \eta (k_{t}^{u}\tilde{g}(\theta_{t}^{u})+ h(\theta_{t}^{u}))$.
\EndFor
\end{algorithmic}
\end{algorithm}
\vspace{-1.5em}

\section{Experiments}

\subsection{Datasets}

\noindent \textbf{CREMA-D}\cite{cao2014crema} is an audio-visual dataset for speech emotion recognition, containing 7,442 video clips of 2-3 seconds from 91 actors speaking several short words. This dataset consists of 6 most usual emotions: \textit{angry}, \textit{happy}, \textit{sad}, \textit{neutral}, \textit{discarding}, \textit{disgust} and \textit{fear}. Categorical emotion labels were collected using crowd-sourcing from 2,443 raters. The whole dataset is randomly divided into 6,698-sample training set and validation set according to the ratio of 9/1, as well as a 744-sample testing set.

\noindent \textbf{Kinetics-Sounds} (KS) \cite{arandjelovic2017look} is a dataset containing 31 human action classes selected from Kinetics dataset~\cite{kay2017kinetics} which contains 400 classes of YouTube videos. All videos are manually annotated for human action using Mechanical Turk and cropped to 10 seconds long around the action. The 31 classes were chosen to be potentially manifested visually and aurally, such as playing various instruments. This dataset contains 19k 10-second video clips (15k training, 1.9k validation, 1.9k test).

\noindent \textbf{VGGSound}\cite{chen2020vggsound} is a large-scale video dataset that contains 309 classes, covering a wide range of audio events in everyday life. All videos in VGGSound are captured ``in the wild'' with audio-visual correspondence in the sense that the sound source is visually evident. The duration of each video is 10 seconds, and the division of the dataset is the same as~\cite{chen2020vggsound}. In our experimental settings, 168,618 videos are used for training and validation, and 13,954 videos are used for testing because some videos are not available now on YouTube.

\noindent \textbf{AVE}\cite{tian2018audio} is an audio-visual video dataset for audio-visual event localization, which covers 28 event classes and consists of 4,143 10-second videos with both auditory and visual tracks as well as frame-level annotations. All videos are collected from YouTube. In experiments, the split of the dataset follows~\cite{tian2018audio}.

\subsection{Experimental settings}
In the experiment, we use the ResNet18-based~\cite{he2016deep} network as the backbones. Specifically, for the visual encoder, we take multiple frames as input, and put them into the 2D network as~\cite{zhao2018sound} does; for the audio encoder, we slightly change the input channel of ResNet18 from 3 to 1 as~\cite{chen2020vggsound} does, and the rest parts keep unchanged. Videos in AVE, Kinetics-Sounds, and  VGGSound lasts 10-second in length and we extract frames with 1 fps. Considering the difference between datasets, 3 frames are uniformly sampled from each 10-second clip as visual inputs. The whole audio data is transformed into a spectrogram of size 257$\times$1,004 by librosa~\cite{mcfee2015librosa} using a window with length of 512 and overlap of 353. For CREMA-D, we extract 1 frame from each of the clip and process audio data into a spectrogram of size 257$\times$299 with window length of 512 and overlap of 353. We use SGD with 0.9 momentum and 1e-4 weight decay as the optimizer. The learning rate is 1e-3 initially and multiplies 0.1 every 70 epochs, and $\alpha$ is determined from $[0,1]$ according to the validation set. 

\begin{table}[t]
\centering
\setlength{\tabcolsep}{3.5mm}{
\begin{tabular}{c|cc|cc}
\toprule[1pt]
Dataset                             & \multicolumn{2}{c|}{CREMA-D}  & \multicolumn{2}{c}{VGGSound}\\ \hline
Method                              & Acc          & mAP& Acc          & mAP          \\ \hline
Audio-only                          & 52.5             & 54.2 & 44.3          & 48.4 \\
Visual-only                         & 41.9             & 43.0 & 31.0          & 34.3\\ \hline
Baseline      & 50.8             & 52.6 & 48.4          & 51.7\\
Concatenation                       & 51.7             & 53.5 & 49.1          & 52.5 \\
Summation                           & 51.5             & 53.5 & 49.1          & 52.4\\
FiLM \cite{perez2018film}            & 50.6             & 52.1& 48.5          & 51.6 \\ \hline
Baseline\dag                           & 54.4             & 56.2        & 50.1          & 53.5\\
Concatenation\dag                   & 61.9             & 63.9       & \textbf{50.6} & \textbf{53.9}\\
Summation\dag                       & \textbf{62.2}    & \textbf{64.3}& 50.4          & 53.6\\

FiLM\dag                            & 55.6             & 57.4& 50.0          & 52.9\\ \bottomrule[1pt]
\end{tabular}}

\caption{Performance on CREMA-D and VGGSound dataset. Combined with OGM-GE, conventional fusion methods consistently gain considerable improvement. \dag~indicates OGM-GE strategy is applied.}
\vspace{-0.5em}
\label{baseline}
\end{table}

\subsection{Comparison on the multimodal task}

\noindent \textbf{Combination with conventional fusion methods.} We first apply OGM-GE in several vanilla fusion methods: baseline, concatenation and summation, and specifically-designed fusion method: FiLM~\cite{perez2018film}, then evaluate the performance on several datasets as shown in Table~\ref{baseline}. Baseline is an introduced vanilla fusion method that uses activation function to mark out the related feature component of one modality compared with another. FiLM~\cite{perez2018film} learns to adaptively influence the output of a neural network by applying an affine transformation to the network's intermediate features. Audio-only and visual-only results are also provided. According to the results, we can find that the performance of each uni-modal model is unbalanced. For example, the audio-only model performance on the VGGSound dataset greatly outperforms its visual-only counterpart. In addition, the accuracy of audio-only model on the CREMA-D dataset is better than all of the vanilla fusion methods, which indicates that the potential of the multimodal model is indeed suppressed. After combining with OGM-GE, the performance of all the vanilla fusion methods consistently gains considerable improvement on different datasets, indicating the effectiveness and satisfactory flexibility of our method.

\begin{table}[h]\small%
\centering
\setlength{\tabcolsep}{3mm}{
\begin{tabular}{c|c|c}
\toprule[1pt]
Dataset                                                & CREMA-D       & KS \\ \hline
Method                                                & Acc          & Acc     \\ \hline
Concatenation                                                & 51.7        & 59.8 \\
Modality-Drop~\cite{feichtenhofer2019slowfast} (audio)                                 & 54.4        & 60.3\\
Modality-Drop~\cite{feichtenhofer2019slowfast} (visual)                            & 53.3        & 61.3\\
Grad-Blending~\cite{wang2020makes}                                  & 56.8       & 62.2\\ \hline
OGM                                           &    59.0      &  61.1\\
OGM-GE                                      & \textbf{61.9}     & \textbf{62.3}\\ \bottomrule[1pt]
\end{tabular}}

\caption{Comparison with other modulation strategies on CREMA-D and Kinetics-Sounds dataset. Modality-Drop~(audio) method casts the audio input to a certain probability, similarly for Modality-Drop~(visual). Compared to concatenation, all other methods make progress, among which our OGM-GE achieves the best performance.}

\label{adjust method}
\end{table}

\noindent \textbf{Comparison with other modulation strategies.} To demonstrate the advantage of OGM-GE, we make comparisons with two main-stream modulation approaches: Modality-Dropout~\cite{xiao2020audiovisual}, and Gradient-Blending~\cite{wang2020makes}. To be fair, the same backbone with concatenation is used in all experiments. Results shown in Table~\ref{adjust method} prove that all the compared methods achieved better performance than the baseline of concatenation more or less, which indicates that the imbalance phenomenon indeed influences the results but also confirms the effectiveness of these methods. We can also notice that our proposed method show the superior performance among all other methods. Firstly, although according to Theorem~\ref{noise_gene} that OGM might theoretically hurt the generalization ability because it would reduce the intensity of the stochastic gradient noise, it is still observed improvements and comparable to other methods, which surprisingly indicates that its positive effect surpasses its negative impact to some extent. Secondly, when boosted with the introduced GE strategy, our method shows the best performance on all the datasets. It should also be noted that Gradient-Blending~\cite{wang2020makes} requires to train additional uni-modal classifiers for modulation, while our method is free of that hence much more efficient and effective in contrast.

\begin{table}[t]
\centering

\setlength{\tabcolsep}{4mm}{
\begin{tabular}{c|c|c}
\toprule[1pt]
Dataset                                           & KS             & VGGSound    \\ \hline
Method                                            &Acc             & Acc            \\ \hline
TSN-AV~\cite{wang2016temporal}                                          & 58.6           & 49.0              \\
TSM-AV~\cite{lin2019tsm}                                           & 60.3           & 48.8                 \\
TBN~\cite{kazakos2019TBN}                                            & 60.8           & 49.4            \\
PSP~\cite{zhou2021positive}                                            & 59.7           & 49.2          \\ \hline
TSN-AV\dag                                       & 59.1           & 49.6             \\
TSM-AV\dag                                       & 62.4           & 49.6                 \\
TBN\dag                                         & \textbf{63.1} & \textbf{50.4}                 \\
PSP\dag                                        & 60.4           & 49.5        \\ \bottomrule[1pt]
\end{tabular}}

\caption{Comparison with existing methods on the Kinetics-Sounds and VGGSound. $\dag$ indicates OGM-GE is applied.}
\label{sota-three}

\end{table}

\noindent \textbf{Combination with existing methods.} For all the datasets, we combine the proposed OGM-GE with several existing methods to further evaluate its flexibility. For Kinetics-Sounds and VGGSound, TSN~\cite{wang2016temporal}, TSM~\cite{lin2019tsm}, TBN~\cite{kazakos2019TBN} and PSP~\cite{zhou2021positive} are compared\footnote{PSP is designed for audio-visual event localization, and here we borrow its backbone to perform the event classification tasks.}. Considering that TSN and TSM are specifically designed to event classification with visual modality, for fair comparison in multimodal setting, we add a ResNet-18 as the audio encoder and only use 3 RGB frames as the visual input, which we refer to as TSN-AV and TSM-AV. In Table~\ref{sota-three} we can find that OGM-GE strengthens these two methods. Moreover, as mentioned before, our method is not limited in the model configurations where the audio and visual encoder have no interaction before fusion process. Concretely, there exist multiple stages of multimodal interaction before the fusion in PSP but OGM-GE can still boost the performance as in other methods, as shown in Table~\ref{sota-three}. Besides, OGM-GE is also combined with x-vector~\cite{pappagari2020x}, i-vector~\cite{heracleous2019comprehensive}, and MWTSM~\cite{ghaleb2019multimodal} on CREMA-D. The results in Table~\ref{sota-CREMA-D} further demonstrate the versatility of OGM-GE.

\label{sec:gene}
\noindent \textbf{Application beyond classification.} 
In order to further validate the versatility of OGM-GE in more general scenarios, we employ it in a representative task, audio-visual event localization, which can be viewed as a fine-grained classification.
The comparison results are shown in Table~\ref{complicated tasks}. We verify the effectiveness of our method on commonly used AVE~\cite{tian2018audio} dataset. Specifically, we insert OGM-GE into two representative frameworks AGVA~\cite{tian2018audio} and PSP~\cite{zhou2021positive}. The improvement brought by OGM-GE demonstrates that the imbalanced phenomenon is universal for different audio-visual architectures and our method still brings consistent performance gain although there exist interactions between different modalities in these AVEL frameworks, which adds difficulty for accurate estimation of the uni-modal performance. Meanwhile, the slight boosts indicate the cross-modal interactions also alleviate the imbalance. Please refer to \emph{Supp. Materials} for more details of the implementation.

\begin{table}[t]
    \begin{center}
    \setlength{\tabcolsep}{8mm}{
    \begin{tabular}{c|c}
    \toprule[1pt]
    Method                                & \multicolumn{1}{c}{Acc} \\ \hline
    I-vector\cite{heracleous2019comprehensive} & 53.6                  \\
    X-vector\cite{pappagari2020x}            & 55.6                   \\
    MWTSM\cite{ghaleb2019multimodal}          & 54.1                  \\\hline
    I-vector\dag                             & 55.3           \\
    X-vector\dag                              & 57.1            \\
    MWTSM\dag                                & \textbf{58.0}            \\\bottomrule[1pt]
    \end{tabular}}
    \vspace{-0.5em}
    \caption{Comparison with the existing methods on CREMA-D dataset. $\dag$ indicates OGM-GE is applied.}
    \label{sota-CREMA-D}
    \vspace{-0.5em}
    \end{center}
\end{table}

\begin{table}[t]
\centering
\setlength{\tabcolsep}{3mm}{
\begin{tabular}{ccc}
\toprule[1pt]
\multicolumn{3}{c}{Audio-visual Event Localization}         \\ \hline
\multicolumn{1}{c|}{w/ or w/o OGM-GE} & w/o & w/          \\ \hline
\multicolumn{1}{c|}{AVGA~\cite{tian2018audio}}   & 72.0    & \textbf{72.8} \\
\multicolumn{1}{c|}{PSP~\cite{zhou2021positive}}    & 76.2    & \textbf{76.9} \\
\bottomrule[1pt]
\end{tabular}
}
\vspace{-0.5em}
\caption{Comparison results on audio-visual event localization between w/ and w/o OGM-GE in two representative frameworks.}
\vspace{-0.5em}
\label{complicated tasks}

\end{table}

\subsection{Ablation study}

\begin{figure}[t]
\centering
    \includegraphics[width=7cm]{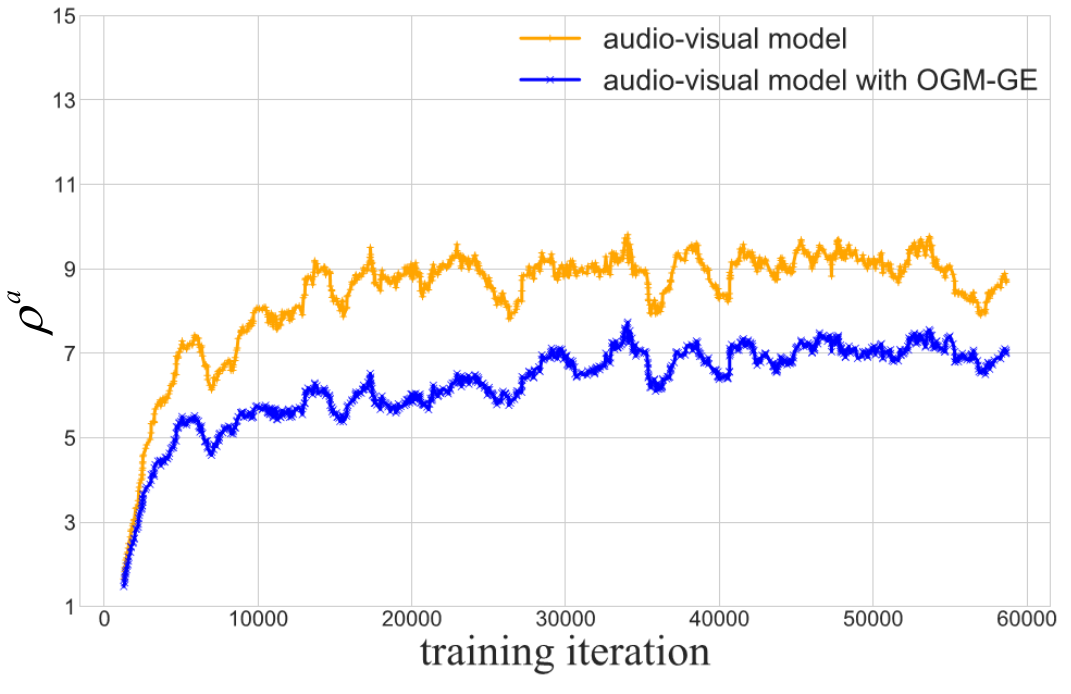}
    \vspace{-1em}
    \caption{Change of discrepancy ratio $\rho^a$ on the VGGSound during training. Our OGM-GE strategy helps the model to have a more balanced optimization.}
    \label{ratio}
    \vspace{-0.5em}
\end{figure}

\noindent \textbf{Imbalance modulation analysis.}
To further analyze OGM-GE, we monitor the change of discrepancy ratio during training. As shown in Figure~\ref{five_lines}, both the performance of audio and visual modality under the multimodal models are improved after the modulation. One interesting observation is that our method performs not as well as others in the beginning but outperforms them in the end. That is because we modulate the gradient and mitigate the optimization of the modality with better performance, by which our method can further exploit the information of the other modalities and finally gain improvement. Moreover, it can be noted from Figure~\ref{ratio} that the discrepancy ratio $\rho^a$ obviously decreases after applied OGM-GE, which is a concrete evidence for the effectiveness of our method. However, it should be noted that the two modalities usually do not have equal contribution for the learning objective because of the natural imbalance existed in various curated datasets, such as VGGSound. This can also be indicated by the results in Table~\ref{baseline} and Figure~\ref{five_lines}.

\noindent \textbf{Adaptation to other optimizer.} To explore the effectiveness of our method in combination with other optimizer, we apply OGM-GE to widely-used Adam optimizer to valid if the OGM-GE strategy works well when optimized by different optimizer. As shown in Table~\ref{optimizer}, combining with the proposed OGM-GE method, the origin SGD and Adam models gain much performance improvement. In addition, based on the same experimental setting, the SGD optimizer obtains better performance than Adam optimizer. The results show that our method can be well adapted to different optimizers, achieving consistent performance improvement.

\noindent \textbf{Analysis of different noise intensities.} As stated in Theorem~\ref{noise_gene}, the intensity of the SGD noise is proportional to the ratio of learning rate to batch-size, and larger noise intensity often leads to better generalization. To empirically validate this theorem, we perform experiments with different batch size and learning rate, using vanilla concatenation fusion method. The introduced GE method, as a noise intensity enhancement strategy, is also solely applied to the vanilla method to explore its effectiveness. As shown in Table~\ref{potential drop}, smaller batch or larger learning rate, i.e. stronger intensity, brings better performance. These results demonstrate the existence of the potential risk of generalization damage when applying OGM, which further validates the necessity of leveraging GE. After introducing extra Gaussian noise, the performance gains considerable improvement, which demonstrates our GE method plays an important role in facilitating the generalization ability.

\begin{table}[t]\small%
\centering
\setlength{\tabcolsep}{3mm}{
\begin{tabular}{c|c|c|c}
\toprule[1pt]
Dataset                   & CREMA-D          & KS                       & VGGSound\\ \hline
Method                    & Acc        & Acc                     & Acc\\ \hline
SGD                         & 51.7     & 59.8                     & 49.1  \\
SGD\dag                     & \textbf{61.9}   & \textbf{63.1}        & \textbf{50.6}\\ \hline
Adam                        & 49.7     & 57.4                       & 47.3   \\
Adam\dag                   & \textbf{54.6}     & \textbf{58.9}    & \textbf{48.2} \\ \bottomrule[1pt]
\end{tabular}}
\vspace{-0.5em}
\caption{Experiments with SGD and Adam optimizers on CREMA-D, Kinetics-Sounds and VGGSound. \dag~indicates OGM-GE is applied.}
\vspace{-0.5em}
\label{optimizer}
\end{table}

\begin{table}[h]
\centering
\setlength{\tabcolsep}{2mm}{
\begin{tabular}{c|c|cl}
\toprule[1pt]
Settings                & CREMA-D     & VGGSound        \\ \hline
(b=64, lr=1e-4) & 50.4         & 48.3            \\
(b=64, lr=5e-4) & 51.0         & 48.7            \\
(b=64, lr=1e-3)  & 51.8        & 49.1 \\ \hline
(b= 64, lr=1e-3)  & 51.8        & 49.1 \\
(b=128, lr=1e-3) & 50.2          & 48.8          \\
(b=256, lr=1e-3)  & 48.6        & 47.7          \\ \hline
(b= 64, lr=1e-3) w/ GE  & 60.2  & 50.3 \\ \bottomrule[1pt]
\end{tabular}}
\vspace{-0.5em}
\caption{Experiments on CREMA-D and VGGSound with different learning rates and batch-size.}
\vspace{-0.5em}
\label{potential drop}
\end{table}

\vspace{-1mm}
\section{Discussion}
In this paper, we propose a simple but effective multimodal learning strategy called On-the-fly Gradient Modulation with Generalization Enhancement (OGM-GE) to alleviate the optimization imbalance problem, facilitating the exploitation of both modalities. This method achieves consistent performance gain on four representative multimodal datasets under various settings and can generally serve as a flexible plug-in strategy for both vanilla fusion method and specifically-designed fusion method as well as existing multimodal models. 

\noindent
\textbf{Limitation.} However, there is still an unresolved issue that, even equipped with OGM-GE, the uni-modal performance in multimodal model still do not surpass the best uni-modal model. We hypothesize that solely leveraging optimization-oriented method could not thoroughly solve the imbalance problem, thus different approaches, such as more advanced fusion strategy or network architectures, must be investigated. We leave this intriguing challenge to future work. Moreover, the versatility that OGM-GE suggests its tremendous potential to more multimodal scenarios, which may include depth, optical flow, language, etc. 

\noindent
\textbf{Broader impacts.} The proposed method is trained on curated datasets that perhaps contains bias, leading to the model inevitably learned such information. This issue is worth further consideration.

\noindent
\textbf{Acknowledgements.} This work is supported by Intelligent Social Governance Platform, Major Innovation \& Planning Interdisciplinary Platform for the “Double-First Class” Initiative, Renmin University of China. This study was also supported by Beijing Outstanding Young Scientist Program (NO.BJJWZYJH012019100020098), the Research Funds of Renmin University of China (NO.21XNLG17), the National Natural Science Foundation of China (NO.62106272), the 2021 Tencent AI Lab Rhino-Bird Focused Research Program (No.JR202141), the Young Elite Scientists Sponsorship Program by CAST, the Large-Scale Pre-Training Program of Beijing Academy of Artificial Intelligence (BAAI) and the Public Computing Cloud, Renmin University of China.

{\small
\bibliographystyle{ieee_fullname}
\bibliography{egbib}

\begin{thebibliography}{10}\itemsep=-1pt

\bibitem{alwassel2020self}
Humam Alwassel, Dhruv Mahajan, Bruno Korbar, Lorenzo Torresani, Bernard Ghanem,
  and Du Tran.
\newblock Self-supervised learning by cross-modal audio-video clustering.
\newblock {\em Advances in Neural Information Processing Systems}, 33, 2020.

\bibitem{antol2015vqa}
Stanislaw Antol, Aishwarya Agrawal, Jiasen Lu, Margaret Mitchell, Dhruv Batra,
  C~Lawrence Zitnick, and Devi Parikh.
\newblock Vqa: Visual question answering.
\newblock In {\em Proceedings of the IEEE international conference on computer
  vision}, pages 2425--2433, 2015.

\bibitem{arandjelovic2017look}
Relja Arandjelovic and Andrew Zisserman.
\newblock Look, listen and learn.
\newblock In {\em Proceedings of the IEEE International Conference on Computer
  Vision}, pages 609--617, 2017.

\bibitem{aytar2016soundnet}
Yusuf Aytar, Carl Vondrick, and Antonio Torralba.
\newblock Soundnet: Learning sound representations from unlabeled video.
\newblock {\em Advances in neural information processing systems}, 29:892--900,
  2016.

\bibitem{cao2014crema}
Houwei Cao, David~G Cooper, Michael~K Keutmann, Ruben~C Gur, Ani Nenkova, and
  Ragini Verma.
\newblock Crema-d: Crowd-sourced emotional multimodal actors dataset.
\newblock {\em IEEE transactions on affective computing}, 5(4):377--390, 2014.

\bibitem{chaudhari2018stochastic}
Pratik Chaudhari and Stefano Soatto.
\newblock Stochastic gradient descent performs variational inference, converges
  to limit cycles for deep networks.
\newblock In {\em 2018 Information Theory and Applications Workshop (ITA)},
  pages 1--10. IEEE, 2018.

\bibitem{chen2020vggsound}
Honglie Chen, Weidi Xie, Andrea Vedaldi, and Andrew Zisserman.
\newblock Vggsound: A large-scale audio-visual dataset.
\newblock In {\em ICASSP 2020-2020 IEEE International Conference on Acoustics,
  Speech and Signal Processing (ICASSP)}, pages 721--725. IEEE, 2020.

\bibitem{du2021improving}
Chenzhuang Du, Tingle Li, Yichen Liu, Zixin Wen, Tianyu Hua, Yue Wang, and Hang
  Zhao.
\newblock Improving multi-modal learning with uni-modal teachers.
\newblock {\em arXiv preprint arXiv:2106.11059}, 2021.

\bibitem{feichtenhofer2019slowfast}
Christoph Feichtenhofer, Haoqi Fan, Jitendra Malik, and Kaiming He.
\newblock Slowfast networks for video recognition.
\newblock In {\em Proceedings of the IEEE/CVF international conference on
  computer vision}, pages 6202--6211, 2019.

\bibitem{gao2020listen}
Ruohan Gao, Tae-Hyun Oh, Kristen Grauman, and Lorenzo Torresani.
\newblock Listen to look: Action recognition by previewing audio.
\newblock In {\em Proceedings of the IEEE/CVF Conference on Computer Vision and
  Pattern Recognition}, pages 10457--10467, 2020.

\bibitem{gazzaniga2006cognitive}
Michael~S Gazzaniga, Richard~B Ivry, and GR Mangun.
\newblock Cognitive neuroscience. the biology of the mind,(2014), 2006.

\bibitem{ghaleb2019multimodal}
Esam Ghaleb, Mirela Popa, and Stylianos Asteriadis.
\newblock Multimodal and temporal perception of audio-visual cues for emotion
  recognition.
\newblock In {\em 2019 8th International Conference on Affective Computing and
  Intelligent Interaction (ACII)}, pages 552--558. IEEE, 2019.

\bibitem{he2019control}
Fengxiang He, Tongliang Liu, and Dacheng Tao.
\newblock Control batch size and learning rate to generalize well: Theoretical
  and empirical evidence.
\newblock {\em Advances in Neural Information Processing Systems},
  32:1143--1152, 2019.

\bibitem{he2016deep}
Kaiming He, Xiangyu Zhang, Shaoqing Ren, and Jian Sun.
\newblock Deep residual learning for image recognition.
\newblock In {\em Proceedings of the IEEE conference on computer vision and
  pattern recognition}, pages 770--778, 2016.

\bibitem{heracleous2019comprehensive}
Panikos Heracleous and Akio Yoneyama.
\newblock A comprehensive study on bilingual and multilingual speech emotion
  recognition using a two-pass classification scheme.
\newblock {\em PloS one}, 14(8):e0220386, 2019.

\bibitem{hu2016temporal}
Di Hu, Xuelong Li, et~al.
\newblock Temporal multimodal learning in audiovisual speech recognition.
\newblock In {\em Proceedings of the IEEE Conference on Computer Vision and
  Pattern Recognition}, pages 3574--3582, 2016.

\bibitem{hu2019deep}
Di Hu, Feiping Nie, and Xuelong Li.
\newblock Deep multimodal clustering for unsupervised audiovisual learning.
\newblock In {\em Proceedings of the IEEE/CVF Conference on Computer Vision and
  Pattern Recognition}, pages 9248--9257, 2019.

\bibitem{hu2019dense}
Di Hu, Chengze Wang, Feiping Nie, and Xuelong Li.
\newblock Dense multimodal fusion for hierarchically joint representation.
\newblock In {\em ICASSP 2019-2019 IEEE International Conference on Acoustics,
  Speech and Signal Processing (ICASSP)}, pages 3941--3945. IEEE, 2019.

\bibitem{ilievski2017multimodal}
Ilija Ilievski and Jiashi Feng.
\newblock Multimodal learning and reasoning for visual question answering.
\newblock In I. Guyon, U.~V. Luxburg, S. Bengio, H. Wallach, R. Fergus, S.
  Vishwanathan, and R. Garnett, editors, {\em Advances in Neural Information
  Processing Systems}, volume~30. Curran Associates, Inc., 2017.

\bibitem{ismail2020improving}
Aya~Abdelsalam Ismail, Mahmudul Hasan, and Faisal Ishtiaq.
\newblock Improving multimodal accuracy through modality pre-training and
  attention.
\newblock {\em arXiv preprint arXiv:2011.06102}, 2020.

\bibitem{jastrzkebski2017three}
Stanislaw Jastrzebski, Zachary Kenton, Devansh Arpit, Nicolas Ballas, Asja
  Fischer, Yoshua Bengio, and Amos Storkey.
\newblock Three factors influencing minima in sgd.
\newblock {\em arXiv preprint arXiv:1711.04623}, 2017.

\bibitem{kay2017kinetics}
Will Kay, Joao Carreira, Karen Simonyan, Brian Zhang, Chloe Hillier, Sudheendra
  Vijayanarasimhan, Fabio Viola, Tim Green, Trevor Back, Paul Natsev, et~al.
\newblock The kinetics human action video dataset.
\newblock {\em arXiv preprint arXiv:1705.06950}, 2017.

\bibitem{kazakos2019epic}
Evangelos Kazakos, Arsha Nagrani, Andrew Zisserman, and Dima Damen.
\newblock Epic-fusion: Audio-visual temporal binding for egocentric action
  recognition.
\newblock In {\em Proceedings of the IEEE/CVF International Conference on
  Computer Vision}, pages 5492--5501, 2019.

\bibitem{kazakos2019TBN}
Evangelos Kazakos, Arsha Nagrani, Andrew Zisserman, and Dima Damen.
\newblock Epic-fusion: Audio-visual temporal binding for egocentric action
  recognition.
\newblock In {\em IEEE/CVF International Conference on Computer Vision (ICCV)},
  2019.

\bibitem{korbar2018cooperative}
Bruno Korbar, Du Tran, and Lorenzo Torresani.
\newblock Cooperative learning of audio and video models from self-supervised
  synchronization.
\newblock {\em arXiv preprint arXiv:1807.00230}, 2018.

\bibitem{lin2019tsm}
Ji Lin, Chuang Gan, and Song Han.
\newblock Tsm: Temporal shift module for efficient video understanding.
\newblock In {\em Proceedings of the IEEE/CVF International Conference on
  Computer Vision}, pages 7083--7093, 2019.

\bibitem{mandt2017stochastic}
Stephan Mandt, Matthew~D Hoffman, and David~M Blei.
\newblock Stochastic gradient descent as approximate bayesian inference.
\newblock {\em arXiv preprint arXiv:1704.04289}, 2017.

\bibitem{mcfee2015librosa}
Brian McFee, Colin Raffel, Dawen Liang, Daniel~PW Ellis, Matt McVicar, Eric
  Battenberg, and Oriol Nieto.
\newblock librosa: Audio and music signal analysis in python.
\newblock In {\em Proceedings of the 14th python in science conference},
  volume~8, pages 18--25. Citeseer, 2015.

\bibitem{nagrani2020speech2action}
Arsha Nagrani, Chen Sun, David Ross, Rahul Sukthankar, Cordelia Schmid, and
  Andrew Zisserman.
\newblock Speech2action: Cross-modal supervision for action recognition.
\newblock In {\em Proceedings of the IEEE/CVF Conference on Computer Vision and
  Pattern Recognition}, pages 10317--10326, 2020.

\bibitem{pappagari2020x}
Raghavendra Pappagari, Tianzi Wang, Jesus Villalba, Nanxin Chen, and Najim
  Dehak.
\newblock x-vectors meet emotions: A study on dependencies between emotion and
  speaker recognition.
\newblock In {\em ICASSP 2020-2020 IEEE International Conference on Acoustics,
  Speech and Signal Processing (ICASSP)}, pages 7169--7173. IEEE, 2020.

\bibitem{parida2020coordinated}
Kranti Parida, Neeraj Matiyali, Tanaya Guha, and Gaurav Sharma.
\newblock Coordinated joint multimodal embeddings for generalized audio-visual
  zero-shot classification and retrieval of videos.
\newblock In {\em Proceedings of the IEEE/CVF Winter Conference on Applications
  of Computer Vision}, pages 3251--3260, 2020.

\bibitem{perez2018film}
Ethan Perez, Florian Strub, Harm De~Vries, Vincent Dumoulin, and Aaron
  Courville.
\newblock Film: Visual reasoning with a general conditioning layer.
\newblock In {\em Proceedings of the AAAI Conference on Artificial
  Intelligence}, volume~32, 2018.

\bibitem{potamianos2004audio}
Gerasimos Potamianos, Chalapathy Neti, Juergen Luettin, and Iain Matthews.
\newblock Audio-visual automatic speech recognition: An overview.
\newblock {\em Issues in visual and audio-visual speech processing}, 22:23,
  2004.

\bibitem{salakhutdinov2009deep}
Ruslan Salakhutdinov and Geoffrey Hinton.
\newblock Deep boltzmann machines.
\newblock In {\em Artificial intelligence and statistics}, pages 448--455.
  PMLR, 2009.

\bibitem{sun2021learning}
Ya Sun, Sijie Mai, and Haifeng Hu.
\newblock Learning to balance the learning rates between various modalities via
  adaptive tracking factor.
\newblock {\em IEEE Signal Processing Letters}, 28:1650--1654, 2021.

\bibitem{tian2018audio}
Yapeng Tian, Jing Shi, Bochen Li, Zhiyao Duan, and Chenliang Xu.
\newblock Audio-visual event localization in unconstrained videos.
\newblock In {\em Proceedings of the European Conference on Computer Vision
  (ECCV)}, pages 247--263, 2018.

\bibitem{wang2020temporal}
Dong Wang, Di Hu, Xingjian Li, and Dejing Dou.
\newblock Temporal relational modeling with self-supervision for action
  segmentation.
\newblock {\em arXiv preprint arXiv:2012.07508}, 2020.

\bibitem{wang2016temporal}
Limin Wang, Yuanjun Xiong, Zhe Wang, Yu Qiao, Dahua Lin, Xiaoou Tang, and Luc
  Van~Gool.
\newblock Temporal segment networks: Towards good practices for deep action
  recognition.
\newblock In {\em European conference on computer vision}, pages 20--36.
  Springer, 2016.

\bibitem{wang2020makes}
Weiyao Wang, Du Tran, and Matt Feiszli.
\newblock What makes training multi-modal classification networks hard?
\newblock In {\em Proceedings of the IEEE/CVF Conference on Computer Vision and
  Pattern Recognition}, pages 12695--12705, 2020.

\bibitem{winterbottom2020modality}
Thomas Winterbottom, Sarah Xiao, Alistair McLean, and Noura~Al Moubayed.
\newblock On modality bias in the tvqa dataset.
\newblock {\em arXiv preprint arXiv:2012.10210}, 2020.

\bibitem{wu2021star}
Bo Wu, Shoubin Yu, Zhenfang Chen, Joshua~B Tenenbaum, and Chuang Gan.
\newblock Star: A benchmark for situated reasoning in real-world videos.
\newblock In {\em Thirty-fifth Conference on Neural Information Processing
  Systems (NeurIPS)}, 2021.

\bibitem{wu2020noisy}
Jingfeng Wu, Wenqing Hu, Haoyi Xiong, Jun Huan, Vladimir Braverman, and
  Zhanxing Zhu.
\newblock On the noisy gradient descent that generalizes as sgd.
\newblock In {\em International Conference on Machine Learning}, pages
  10367--10376. PMLR, 2020.

\bibitem{xiao2020audiovisual}
Fanyi Xiao, Yong~Jae Lee, Kristen Grauman, Jitendra Malik, and Christoph
  Feichtenhofer.
\newblock Audiovisual slowfast networks for video recognition.
\newblock {\em arXiv preprint arXiv:2001.08740}, 2020.

\bibitem{xie2021artificial}
Zeke Xie, Fengxiang He, Shaopeng Fu, Issei Sato, Dacheng Tao, and Masashi
  Sugiyama.
\newblock Artificial neural variability for deep learning: On overfitting,
  noise memorization, and catastrophic forgetting.
\newblock {\em Neural computation}, 33(8):2163--2192, 2021.

\bibitem{zhao2018sound}
Hang Zhao, Chuang Gan, Andrew Rouditchenko, Carl Vondrick, Josh McDermott, and
  Antonio Torralba.
\newblock The sound of pixels.
\newblock In {\em Proceedings of the European conference on computer vision
  (ECCV)}, pages 570--586, 2018.

\bibitem{zhou2021positive}
Jinxing Zhou, Liang Zheng, Yiran Zhong, Shijie Hao, and Meng Wang.
\newblock Positive sample propagation along the audio-visual event line.
\newblock In {\em Proceedings of the IEEE/CVF Conference on Computer Vision and
  Pattern Recognition}, pages 8436--8444, 2021.

\bibitem{zhou2019toward}
Mo Zhou, Tianyi Liu, Yan Li, Dachao Lin, Enlu Zhou, and Tuo Zhao.
\newblock Toward understanding the importance of noise in training neural
  networks.
\newblock In {\em International Conference on Machine Learning}, pages
  7594--7602. PMLR, 2019.

\bibitem{zhu2018anisotropic}
Zhanxing Zhu, Jingfeng Wu, Bing Yu, Lei Wu, and Jinwen Ma.
\newblock The anisotropic noise in stochastic gradient descent: Its behavior of
  escaping from minima and regularization effects.
\newblock {\em stat}, 1050:21, 2018.

\end{thebibliography}
}

\end{document}